\documentclass{article}

\PassOptionsToPackage{numbers, compress}{natbib}

     \usepackage[preprint]{neurips_2022}

\usepackage[utf8]{inputenc} 
\usepackage[T1]{fontenc}    
\usepackage{hyperref}       
\usepackage{url}            
\usepackage{booktabs}       
\usepackage{amsfonts}       
\usepackage{nicefrac}       
\usepackage{microtype}      
\usepackage{xcolor}         
\usepackage{upgreek}
\usepackage{enumerate}
\usepackage{amsmath}
\usepackage{amssymb}
\usepackage{relsize}
\usepackage{graphicx}

\newcommand{\bs}{\boldsymbol}
\newcommand{\tf}{\textbf}

\newcommand{\tet}{\textit}

\DeclareMathOperator*{\Min}{argmin}

\title{scICML: Information-theoretic Co-clustering-based Multi-view Learning for the Integrative Analysis of Single-cell Multi-omics data}

\author{%
  Pengcheng Zeng\thanks{Corresponding author. Email: \texttt{zengpch@shanghaitech.edu.cn}}\\
  Institute of Mathematical Sciences\\
  ShanghaiTech University\\
  Shanghai, China \\
  \AND
   Zhixiang Lin\\
  Department of Statistics \\
  The Chinese University of Hong Kong\\
  Hong Kong, China SAR\\
}

\begin{document}

\maketitle

\begin{abstract}
Modern high-throughput sequencing technologies have enabled us to profile multiple molecular modalities from the same single cell, providing unprecedented opportunities to assay celluar heterogeneity from multiple biological layers. However, the datasets generated from these technologies tend to have high level of noise and are highly sparse, bringing challenges to data analysis. In this paper, we develop a novel information-theoretic co-clustering-based multi-view learning (scICML) method for multi-omics single-cell data integration. scICML utilizes co-clusterings to aggregate similar features for each view of data and uncover the common clustering pattern for cells. In addition, scICML automatically matches the clusters of the linked features across different data types for considering the biological dependency structure across different types of genomic features. Our experiments on four real-world datasets demonstrate that scICML improves the overall clustering performance and provides biological insights into the data analysis of peripheral blood mononuclear cells.
\end{abstract}

\section{Introduction}
Modern technological advances have enabled us to profile rich high-dimensional single-cell multi-omics data at an unprecedented resolution, and the available protocols and multi-omics datasets are growing at an increasing pace. These protocols include M\&T-seq \citep{Angermueller2016}, scNMT-seq \citep{Clark2018}, sci-CAR-seq \citep{Cao2018}, scCAT-seq \citep{Liu2019}, Paired-seq \citep{Zhu2019}, SNARE-seq \citep{Chen2019}, SHARE-seq \citep{Ma2020} and Paired-Tag \citep{Zhu2021}. The resulting single-cell multi-omic datsets can provide insights into the cell's phenotype and links to its genotype \citep{Macau2017}. However, such data tend to have high level of noise and are highly sparse \citep{CT2018}. For example, there are over 99\% zeros in scATAC-seq data in sci-CAR-seq \citep{Cao2018}. These characteristics bring the challenge of analyzing the multi-omics single-cell data. To analyze the complex biological process varying across cells, we need to integrate different types of genomic features via flexible but rigorous computational methods. The methods of data integration are growing recently, and most of them are designed for data measured in different cells and sampled from the same cell population. These methods include Seurat (versions 2 and 3) \citep{Butler2018, Stuart2019}, MOFA \citep{Arge2018}, \tet{couple}NMF \citep{Duren2018}, scVDMC \citep{Zhang2018}, LIGER \citep{Welch2019}, scACE \citep{Lin2019},  \tet{couple}CoC \citep{Zeng2020}, \tet{couple}CoC+ \citep{Zeng2021}, scMC \citep{Zhang2021} and scAMACE \citep{Wangwu2021}. A more comprehensive discussion on integration of single-cell genomic data is presented in \cite{David2020}. Only a few computational methods, including MOFA+ \citep{Arge2020}, scAI \citep{Jin2020}, Seurat (version 4) \citep{Hao2021} and JSNMF \citep{Ma2022},  have been developed for integrative analysis of single-cell genomic data that are derived from the same cells.

While integrating single-cell multi-omics data, we should take into account an important issue - how to link data from multiple omics in a way that is biologically meaningful? As an example, consider the setting where scRNA-seq and scATAC-seq are profiled on the same cells. A subset of features in scATAC-seq data are linked with scRNA-seq data, because promoter accessibility/gene activity score are directly linked with gene expression. Effectively connecting the linked features across data types is expected to be helpful in the integrative analysis of multi-omics data. However, none of the aforementioned integrative methods have considered this issue. To deal with it effectively, we present an information-theoretic co-clustering-based multi-view learning (scICML) method for the integrative analysis of multi-omics data measured from the same cells.

The presented work contains three major contributions: 1) We propose a novel multi-view learning framework, scICML, that can aggregate similar features for each view of data and uncover the common clustering pattern for cells. 2) It can automatically match the clusters of the linked features across data types, which takes into account the biological dependency across different types of genomic features. 3) It can provide biological interpretation for the aggregating pattern of cells for well-studied data such as peripheral blood mononuclear cells.

\section{Method}
In this section, we first introduce the information-theoretic co-clustering framework for a single view of data \cite{Dhillon2003}, and then extend it to our framework of information-theoretic co-clustering-based multi-view learning for single-cell multi-omics data. We treat scATAC-seq data and scRNA-seq data that are measured from the same cells as two views of the single-cell genomic data. We assume that a subset of the features, i.e., gene activity score in scATAC-seq data are linked with the gene expression in scRNA-seq data; and the other features, i.e., accessibility of distal peaks in scATAC-seq data, are not directly linked with the genes in scRNA-seq data. We expect to improve the clustering performance of the cells by (a) integrating these views of data for uncovering the common shared aggregating pattern for cells; and (b) matching the aggregating pattern of linked features across data types for considering the biological dependency across different types of genomic features.

\subsection{Information-theoretic co-clustering}
\label{sub:itcc}
Let $D$ be a $q$ by $n$ matrix representing a dataset with $q$ features for $n$ cells. Let $X$ and $Y$ be discrete random variables, taking values from the sets $\{1,2,\dots,q\}$ and $\{1,2,\dots,n\}$, respectively.  $X$ represents the possible outcome of feature labels and $Y$ represents the possible outcome of cell labels. Let $p(X,Y)$ be the joint probability distribution for $X$ and $Y$, and define $p(X = x, Y = y)$ as the probability of the $x$-th feature being active in the $y$-th cell: the more active the feature, the higher the value. $p(X,Y)_{q \times n}$ is estimated from the normalized dataset, i.e. scaling the data matrix $D$ to have total sums equal to 1, and we have
\begin{equation}
\label{p}
p(X = x, Y = y)=\frac{D_{xy}}{\sum_{i=1}^{q}\sum_{j=1}^{n}{D_{ij}}},
\end{equation}
where $x\in\{1,\dots,q\}, y\in\{1,\dots,n\}$. Co-clustering aims at clustering similar features into clusters and similar cells into clusters. Suppose that we want to cluster the features into $K$ clusters, and the cells into $N$ clusters. We denote the clusters of features and cells as the possible outcomes of discrete random variables $\tilde{X}$ and $\tilde{Y}$, where $\tilde{X}$ and $\tilde{Y}$ take values from the sets of feature cluster indexes $\{1,\dots,K\}$ and cell cluster indexes $\{1,\dots,N\}$, respectively. To map features to feature clusters and cells to cell clusters, we use $C_{X}(\cdot)$ and $C_{Y}(\cdot)$ to represent the clustering functions of features and cells, respectively. $C_{X}(x) = \tilde{x}$ ($\tilde{x} \in \{ 1,\dots,K\}$) indicates that the feature $x$ belongs to the cluster $\tilde{x}$, and $C_{Y}(y) = \tilde{y}$ ($\tilde{y} \in \{ 1,\dots,N\}$) indicates that the cell $y$ belongs to the cluster $\tilde{y}$. We then let $\tilde{p}(\tilde{X},\tilde{Y})$ be the joint probability distribution for $\tilde{X}$ and $\tilde{Y}$, and this distribution can be expressed as
\begin{equation}
\label{prob}
\tilde{p}(\tilde{X}= \tilde{x}, \tilde{Y} = \tilde{y}) = \sum_{x \in \{C_{X}(x) = \tilde{x}\}}\sum_{y \in \{C_{Y}(y) = \tilde{y}\}}p(X =x,Y=y).
\end{equation}
Note that $\tilde{p}(\tilde{X} = \tilde{x}, \tilde{Y} = \tilde{y})$ is connected to $p(X,Y)$ via the clustering functions $C_{X}(\cdot)$ and $C_{Y}(\cdot)$. The matrix $\tilde{p}(\tilde{X},\tilde{Y})_{K\times N}$ can be interpreted as the low-dimensional representations for the feature clusters and cell clusters in the data $D$. The goal of information-theoretic co-clustering \cite{Dhillon2003} is to find the optimal clustering functions $C_{X}(\cdot)$ and $C_{Y}(\cdot)$ to minimize the loss of mutual information:
\begin{equation}
\label{eq:lossD}
\ell(C_X,C_Y) = I(X;Y) - I(\tilde{X};\tilde{Y}),
\end{equation}
where $I(\cdot)$ denotes the function of mutual information, and we have $I(X;Y) = \sum_{x}\sum_{y}p(x,y)\text{log}\frac{p(x,y)}{p(x)p(y)}$, and
$I(\tilde{X};\tilde{Y}) = \sum_{\tilde{x}}\sum_{\tilde{y}}\tilde{p}(\tilde{x},\tilde{y})\text{log}\frac{\tilde{p}(\tilde{x},\tilde{y})}{\tilde{p}(\tilde{x})\tilde{p}(\tilde{y})}$.

\subsection{The framework of scICML}
\label{framework}
We now extend the framework of information-theoretic co-clustering \cite{Dhillon2003} to multiple data matrices, and simultaneously perform matching of the feature clusters across datasets. Here we set $v=1$ and 2 to represent the parallel scRNA-seq data and scATAC-seq data, respectively. The gene activity score in scATAC-seq data is linked with gene expression in scRNA-seq data and the accessibility of distal peaks in scATAC-seq data is not directly linked with the genes in scRNA-seq data. We further use index symbols $(1,v)$ and $(2,v)$ to represent the linked part and unlinked part of data in the $v$th view, respectively.  We write the $v$th view of data as $D^{(v)} = \begin{bmatrix}
    D^{(1,v)} \\
    D^{(2,v)}
\end{bmatrix}_{q^{(v)} \times n}$, where $q^{(v)} = q^{(1,v)}+q^{(2,v)}$, representing the sum of the number of features in the linked matrix $ D^{(1,v)}$ and that in the unlinked matrix $D^{(2,v)}$. Here we assume that $[D^{(1,1)}]_{q^{(1,1)} \times n}$ (i.e., gene expression) is directly linked with $[D^{(1,2)}]_{q^{(1,2)} \times n}$ (i.e., gene activity score), and $[D^{(2,1)}]_{q^{(2,1)} \times n}$ (i.e., gene expression) is not directly linked with  $[D^{(2,2)}]_{q^{(2,2)} \times n}$ (i.e., accessibility of distal peaks). We consider these submatrices $D^{(1,1)}, D^{(1,2)}, D^{(2,1)}, D^{(2,2)}$ as the four views of single-cell genomic data.

Similar to definitions of information-theoretic co-clustering for any single data matrix $D$ in Section \ref{sub:itcc}, we have the loss of mutual information for co-clustering the data $D^{(l,v)}$:
\begin{equation}
\label{eq:lossDij}
\ell^{(l,v)}(C_{X^{(l,v)}},C_Y) = I(X^{(l,v)};Y) - I(\tilde{X}^{(l,v)};\tilde{Y}),\hspace{3mm}l=1,2;\hspace{1mm}v=1,2,
\end{equation}
where $X^{(l,v)}$ is the discrete random variable representing the feature labels in data $D^{(l,v)}$. We have $I(X^{(l,v)};Y) = \sum_{x}\sum_{y}p^{(l,v)}(x,y)\text{log}\frac{p^{(l,v)}(x,y)}{p^{(l,v)}(x)p^{(l,v)}(y)}$, where $x\in\{1,\dots,q^{(l,v)}\}, y\in\{1,\dots,n\}$. $\tilde{X}^{(l,v)}$ is the discrete random variables representing the feature cluster labels in data $D^{(l,v)}$. We have $I(\tilde{X}^{(l,v)};\tilde{Y}) = \sum_{\tilde{x}}\sum_{\tilde{y}}\tilde{p}^{(l,v)}(\tilde{x},\tilde{y})\text{log}\frac{\tilde{p}^{(l,v)}(\tilde{x},\tilde{y})}{\tilde{p}^{(l,v)}(\tilde{x})\tilde{p}^{(l,v)}(\tilde{y})}$, where $\tilde{x}\in\{1,\dots,K^{(l,v)}\}, \tilde{y}\in\{1,\dots,N\}$. $C_{X^{(l,v)}}$ is the clustering function for the features in data $D^{(l,v)}$. Note that since the cells in data $\{D^{(l,v)}: l=1,2;v=1,2\}$ are the same, these four data matrices share the same clustering function $C_Y$ for cells. The function $C_Y$ is the key for multi-view learning among the datasets, and it helps to cluster cells by taking advantage of information from the four views of data.

Because the gene expression data $D^{(1,1)}$ is assumed directly linked with the gene activity score data $D^{(1,2)}$, the clustering functions $C_{X^{(1,1)}}$ and $C_{X^{(1,2)}}$ for the linked features are assumed to be similar. The feature clusters in $D^{(1,1)}$ and $D^{(1,2)}$ should be matched, representing similar aggregating patterns for features across the two datasets. Denote $h$ as a permutation of size $K^{(1,1)}$ for the indexes of the feature clusters in data $D^{(1,1)}$ (here we set $K^{(1,1)} = K^{(1,2)}$, i.e., the numbers of feature clusters are equal in data $D^{(1,1)}$ and $D^{(1,2)}$). We then use $D_{\text{KL}} (\tilde{p}^{(1,1)}(\tilde{X}^{(1,1)}_{h}, \tilde{Y}) || \tilde{p}^{(1,2)}(\tilde{X}^{(1,2)}, \tilde{Y}))$ to measure the statistical distance between two probability distributions $\tilde{p}^{(1,1)}(\tilde{X}^{(1,1)}_{h}, \tilde{Y})$ and $\tilde{p}^{(1,2)}(\tilde{X}^{(1,2)}, \tilde{Y})$, where $D_{\text{KL}}(\cdot || \cdot)$ is Kullback-Leibler ($\text{KL}$) divergence \cite{Cover1991}. We obtain $\tilde{p}^{(1,1)}(\tilde{X}^{(1,1)}_{h}, \tilde{Y})$ by adjusting the order of rows of $\tilde{p}^{(1,1)}(\tilde{X}^{(1,1)}, \tilde{Y})$ based on $h$. The smaller the KL divergence, the more similar the feature clusters adjusted by $h$.

To co-cluster the four views of data simultaneously, and to match feature clusters across the data $D^{(1,1)}$ and $D^{(1,2)}$, we propose the following optimization problem in scICML:
\begin{equation}
\label{obj}
\Min_{C_{Y},\{C_{X^{(l,v)}}\}_{l,v=1}^{2},h}\sum\limits_{l=1}^{2}\sum\limits_{v=1}^{2}\ell^{(l,v)}(C_{X^{(l,v)}},C_Y)+\alpha D_{\text{KL}} (\tilde{p}^{(1,1)}(\tilde{X}^{(1,1)}_{h}, \tilde{Y}) || \tilde{p}^{(1,2)}(\tilde{X}^{(1,2)}, \tilde{Y})).
\end{equation}
Aggregating similar features for each view of data reduces the noise in the single-cell data, and utilizing the information of the same cell across the four views of data can generally boost the clustering performance of cells. The term $D_{\text{KL}} (\tilde{p}^{(1,1)}(\tilde{X}^{(1,1)}_{h}, \tilde{Y}) || \tilde{p}^{(1,2)}(\tilde{X}^{(1,2)}, \tilde{Y}))$ borrows information from the linked data $D^{(1,1)}$ and $D^{(1,2)}$ by each other, thereby further improving the multi-omics data integration. $\alpha$ is a hyperparameter that controls the contribution of feature clusters matching across $D^{(1,1)}$ and $D^{(1,2)}$. Our framework has five hyperparameters: the number of cell clusters $N$, the numbers of feature clusters $\{K^{(l,v)}\}_{l,v=1}^{2}$(where $K^{(1,1)} = K^{(1,2)}$), and $\alpha$. The way of choosing these hyperparameters will be discussed in Section \ref{hypersele}.

\subsection{The optimization for scICML}
We first reformulate the loss in mutual information in Equation (\ref{eq:lossD}) into the form of KL divergence \cite{Dhillon2003,Dai2008}:
\begin{equation}
\label{obj2}
\ell(C_{X},C_{Y}) = D_{\text{KL}}(p(X,Y)||p^{\ast}(X,Y)),
\end{equation}
where $p^{\ast}(X,Y)$ is defined as
\begin{equation}
\label{dis3}
p^{\ast}(X = x, Y = y)
             \triangleq p(\tilde{X} = C_{X}(x), \tilde{Y} = C_{Y}(y))\times  \frac{p(X = x)}{p(\tilde{X} = C_{X}(x))} \times \frac{p(Y = y)}{p(\tilde{Y} = C_{Y}(y))}.
\end{equation}
Further, we have
\begin{equation}
\label{obj3}
\resizebox{\hsize}{!}{$
\begin{split}
D_{\text{KL}}(p(X,Y)||p^{\ast}(X,Y))  = & \sum_{i = 1}^{K}\sum_{x \in \{x: C_{X}(x) = i\}}p(X = x)D_{\text{KL}}(p(Y|X = x)||p^{\ast}(Y|\tilde{X} = i, X = x))\\
                           =&\sum_{j = 1}^{N}\sum_{y \in \{y: C_{Y}(y) = j\}}p(Y = y)D_{\text{KL}}(p(X|Y = y)||p^{\ast}(X|\tilde{Y} = j, Y = y)),
\end{split}
$}
\end{equation}
where $p^{\ast}(Y = y|\tilde{X} = i, X = x) \triangleq \frac{p^{\ast}(X = x, Y = y)}{p(X = x)}$, $y = 1,\dots, n$, for any $x \in \{x: C_{X}(x) = i\}$, and $p^{\ast}(X = x|\tilde{Y} = j, Y = y) \triangleq \frac{p^{\ast}(X = x, Y = y)}{p(Y = y)}$, $x = 1,\dots, q$, for any $y \in \{y: C_{Y}(y) = j\}$. Details on the derivation for formulas in Equations (\ref{obj2}) and (\ref{obj3}) are presented in \cite{Dhillon2003} and \cite{Dai2008}.

Now we can rewrite optimization problem in Equation (\ref{obj}) as:
\begin{equation}
\label{obj4}
\resizebox{\hsize}{!}{$
\begin{split}
\Min_{C_{Y},\{C_{X^{(l,v)}}\}_{l,v=1}^{2},h}\sum\limits_{l=1}^{2}\sum\limits_{v=1}^{2}D_{\text{KL}}(p^{(l,v)}(X^{(l,v)},Y)||p^{(l,v)\ast}(X^{(l,v)},Y)+\alpha D_{\text{KL}} (\tilde{p}^{(1,1)}(\tilde{X}^{(1,1)}_{h}, \tilde{Y}) || \tilde{p}^{(1,2)}(\tilde{X}^{(1,2)}, \tilde{Y})).
\end{split}
$}
\end{equation}

We next solve the optimization problem in Equation (\ref{obj4}) by iteratively updating  $C_{Y}$, $\{C_{X^{(l,v)}}\}_{l,v=1}^{2}$ and $h$ as follows:

\tf{Step 1: given $C_{Y}$ and $h$, update $\{C_{X^{(l,v)}}\}_{l,v=1}^{2}$}. The optimization problem in Equation (\ref{obj4}) is equivalent to minimizing
\[
\sum_{i = 1}^{K^{(l,v)}}\sum_{x \in \{x: C_{X^{(l,v)}}(x) = i\}}p^{(l,v)}(X^{(l,v)} = x)\times U^{(l,v)}(\tilde{X}^{(l,v)} = i, X^{(l,v)} = x),
\]
where
\begin{equation}
\resizebox{\textwidth}{!}{$
\begin{split}
U^{(l,v)}(\tilde{X}^{(l,v)} = i, X^{(l,v)} = x) \triangleq & D_{\text{KL}}(p^{(l,v)}(Y|X^{(l,v)} = x)||p^{^{(l,v)\ast}}(Y|\tilde{X}^{(l,v)} = i, X^{(l,v)} = x))\\ & +\frac{\alpha (2-l)D_{\text{KL}} (\tilde{p}^{(1,1)}(\tilde{X}^{(1,1)}_{h}, \tilde{Y}) || \tilde{p}^{(1,2)}(\tilde{X}^{(1,2)}, \tilde{Y}))}{q^{(l,v)}\times p^{(l,v)}(X^{(l,v)}=x)}.
\end{split}
$}
\end{equation}
We iteratively update the cluster assignment $C_{X^{(l,v)}}(x)$ for each feature $x$ ($x = 1,\dots, q^{(l,v)}$) in the data $D^{(l,v)}$, fixing the cluster assignment for the other features:
\begin{equation}
\label{cx}
C_{X^{(l,v)}}(x) = \Min_{i \in \{1,\dots,K^{(l,v)}\}}U^{(l,v)}(\tilde{X}^{(l,v)} = i, X^{(l,v)} = x)).
\end{equation}

\tf{Step 2: given $\{C_{X^{(l,v)}}\}_{l,v=1}^{2}$ and $h$, update $C_{Y}$}. The optimization problem in Equation (\ref{obj4}) is equivalent to minimizing
\[
\sum_{j = 1}^{N}\sum_{y \in \{y: C_{Y}(y) = j\}}W(\tilde{Y} = j, Y = y),
\]
where
\begin{equation}
\resizebox{\textwidth}{!}{$
\begin{split}
W(\tilde{Y} = j, Y = y) \triangleq & \sum\limits_{l=1}^{2}\sum\limits_{v=1}^{2}p^{(l,v)}(Y=y)D_{\text{KL}}(p^{(l,v)}(X^{(l,v)}|Y=y)||p^{^{(l,v)\ast}}(X^{(l,v)}|\tilde{Y} = j, Y = y))\\ & +\frac{\alpha D_{\text{KL}} (\tilde{p}^{(1,1)}(\tilde{X}^{(1,1)}_{h}, \tilde{Y}) || \tilde{p}^{(1,2)}(\tilde{X}^{(1,2)}, \tilde{Y}))}{n}.
\end{split}
$}
\end{equation}
We iteratively update the cluster assignment $C_{Y}(y)$ for each cell $y$ ($y = 1,\dots, n$), fixing the cluster assignment for the other cells:
\begin{equation}
\label{cy}
C_{Y}(y) = \Min_{j \in \{1,\dots,N\}}W(\tilde{Y} = j, Y = y).
\end{equation}

\tf{Step 3: given $\{C_{X^{(l,v)}}\}_{l,v=1}^{2}$ and $C_{Y}$, update $h$}. The optimization problem in Equation (\ref{obj4}) is equivalent to minimizing
$D_{\text{KL}} (\tilde{p}^{(1,1)}(\tilde{X}^{(1,1)}_{h}, \tilde{Y}) || \tilde{p}^{(1,2)}(\tilde{X}^{(1,2)}, \tilde{Y}))$. Thus,
\begin{equation}
\label{h}
h = \Min_{h}D_{\text{KL}} (\tilde{p}^{(1,1)}(\tilde{X}^{(1,1)}_{h}, \tilde{Y}) || \tilde{p}^{(1,2)}(\tilde{X}^{(1,2)}, \tilde{Y})).
\end{equation}
Note that there are $K^{(1,1)}!$ total of combinations of $h$. For each given $h$, the way of computing this KL divergence has been described in Section \ref{framework}.

To summarize, the procedures of the scICML algorithm are as follows:
\begin{enumerate}
\item Initialize the clustering functions $\{C^{[0]}_{X^{(l,v)}}\}_{l,v=1}^{2}$ and $C^{[0]}_{Y}$ first, and then initialize $h^{[0]}$ based on Equation (\ref{h}), and initialize $\{p^{(l,v)\ast[0]}(X^{(l,v)},Y)\}_{l,v=1}^{2}$ based on Equation (\ref{dis3}).
\item Iterate (a)-(b) until convergence: \\
(a). Fix $\{p^{(l,v)\ast[t-1]}(X^{(l,v)},Y)\}_{l,v=1}^{2}$ and $h^{[t-1]}$, and sequentially update  $\{C^{[t]}_{X^{(l,v)}}\}_{l,v=1}^{2}$, $C^{[t]}_{Y}$ and $h^{[t]}$ based on the Equations (\ref{cx}), (\ref{cy}) and (\ref{h}), respectively.\\
(b). Fix $\{C^{[t]}_{X^{(l,v)}}\}_{l,v=1}^{2}$ and $C^{[t]}_{Y}$, and update $\{p^{(l,v)\ast[t]}(X^{(l,v)},Y)\}_{l,v=1}^{2}$  based on Equation (\ref{dis3}).
\item Output the clustering result $C_{Y}$ in the last iteration.
\end{enumerate}

The objective function in Equation (\ref{obj4}) is non-increasing in the updates of $\{C_{X^{(l,v)}}\}_{l,v=1}^{2}$ (Equation (\ref{cx})), $C_{Y}$ (Equation (\ref{cy})) and $h$ (Equation (\ref{h})), and the algorithm will converge to a local minimum. Finding the global optimal solution is NP-hard. Additionally, the algorithm converges in a finite number of iterations due to the finite search space. We stop the iterations when the difference of the loss function (Equation  (\ref{obj4})) is less than $10^{-4}$ between two adjacent iterations. In practice, this algorithm works well in single-cell multi-omics data integration.

\section{Related works}
There are several information-theoretic co-clustering-based methods, including STC \citep{Dai2008}, \tet{couple}CoC \citep{Zeng2020} and \tet{couple}CoC+ \citep{Zeng2021}. All these methods aim at clustering the target data by transferring knowledge from the auxiliary data or source data. Our method scICML differs from them in three aspects: (a) STC is mainly used for tasks in image clustering or text mining. \tet{couple}CoC and \tet{couple}CoC+ are designed to integrate multiple single-cell genomic data derived from different cells. Our method scICML is designed to integrate single-cell multi-omics data derived from the same cells. (b) In their algorithms, two or three co-clusterings are performed simultaneously on the target data and the auxiliary data or source data to uncover the common shared feature clusters. In scICML, we simultaneously perform co-clusterings on four views of data to uncover the common shared sample/cell clusters. (c) Both \tet{couple}CoC and \tet{couple}CoC+ perform matching of cell types from different cells while our method scICML performs matching of feature clusters from the same cells.

In recent years, several computational methods for single-cell multi-omics data from the same cells have been developed. scAI \citep{Jin2020} aggregates epigenomic data in cell subpopulations that exhibit similar gene expression and epigenomic profiles through iterative learning in an unsupervised manner. MOFA+ \citep{Arge2020} uses computationally efficient variational inference to reconstruct a low-dimensional representation of the data and allows to model variation across multi-omics single-cell genomic data. Seurat (version 4) \citep{Hao2021} introduces the ``weighted nearest neighbor'' analysis to learn the relative utility of each genomic feature in each cell, enabling an integrative analysis of multi-omics data.  Our method scICML is distinct from these methods in two aspects: First, these methods are based on non-negative matrix factorization (scAI), probabilistic generative model (MOFA+) and weighted nearest neighbor graph (Seurat V4), respectively, while scICML is based on information-theoretic co-clustering. Second, among these methods only scICML considers the biological dependency across different types of genomic features across multi-omics data.

\section{Experiments}
\label{expe}
We evaluate scICML on four real datasets of single cell multiome ATAC and gene expression. We compare our method with four baselines, including scAI \citep{Jin2020}, MOFA+ \citep{Arge2020}, Seurat V4 \citep{Hao2021} and scICML$_{0}$. We note that scICML$_{0}$ is a simplified version of scICML, where we set $\alpha$ in Equation (\ref{obj}) as 0. We evaluate scICML$_{0}$ to check whether the matching of feature clusters across the linked data can improve the clustering performance. We implement scICML on MATLAB.\footnote{All the code and preprocessed data are included in the supplementary material.}

\subsection{Data description}
\label{des}
The four real datasets of the paired single cell multiome ATAC and gene expression include peripheral blood mononuclear cells from two healthy donors - granulocytes removed through cell sorting (one dataset has 2,711 cells - dataset 1 and another one has 11,898 cells - dataset 2), frozen human healthy brain tissue (cerebellum) (3,233 cells - dataset 3) and fresh cortex, hippocampus, and ventricular zone from embryonic mouse brain (E18) (4,878 cells - dataset 4). All the datasets are processed by Cell Ranger ARC 2.0.0 and are available at the 10X Genomics website(\url{https://www.10xgenomics.com/}). The numbers of cell clusters of gene expression data in datasets 1-4 are given as 9, 17, 8 and 10, respectively, by the secondary analysis outputs for the data at the 10X Genomics website.

\subsection{Data preprocessing and feature selection}
To obtain the linked data, we first compute the gene activity score using the gene scoring approach by \cite{Chen2019}: the distance-weighted sum of reads (peak values in scATAC-seq data) within or near the region gives the accessibility at each transcription start site (TSS). Second, we extract the set of genes that have both gene expression and gene activity score, which can be considered as the linked features. Third, we choose 1,000 most highly variable genes from this set using the R toolkit Seurat \citep{Butler2018, Stuart2019}. We then obtain the linked data $[X^{(1,1)}]_{1,000 \times n}$ and $[X^{(1,2)}]_{1,000 \times n}$. We also use the R toolkit Seurat to choose 1,000 most highly variable genes from the set of genes that are not included in the linked data, and we have the unlinked scRNA-seq data $[X^{(2,1)}]_{1,000 \times n}$. We choose 1,000 unlinked features in scATAC-seq data, corresponding to the top 1,000 largest summation of peak values over all cells, and we have the unlinked scATAC-seq data $[X^{(2,2)}]_{1,000 \times n}$. We take log transformation for scRNA-seq data to alleviate the effect of extreme values in the data matrices. scICML can automatically adjust for sequencing depth, so we do not need to normalize for sequencing depth.

\subsection{Experimental details and evaluation metrics}
\label{hypersele}
For the selection of hyperparameters in scICML, we can pre-determine the number of cell clusters $N$ using the Calinski-Harabasz (CH) index \cite{Calin1974}, and choose the numbers of feature clusters $\{K^{(l,v)}\}_{l,v=1}^{2}$ and the value of $\alpha$ using the approach similar to \tet{couple}CoC+ \citep{Zeng2021}. For all datasets in this work, we set the numbers of cell clusters $N$ the same as that given by the secondary analysis outputs at the 10X Genomics website, set $K^{(1,1)}=K^{(1,2)}=K^{(2,1)}=K^{(2,2)}=10$ for the numbers of feature clusters, and set $\alpha=1$. This selection of hyperparamters works well in real data analysis. The baseline Seurat V4 produces cluster memberships of cells by itself. We implement Louvain clustering \cite{chen2019asse} after dimension reduction by the baselines scAI and MOFA+. When computing the cluster memberships of cells in four real datasets by all baselines except Seurat V4, we set the numbers of cell clusters the same as that given by the secondary analysis outputs at the 10X Genomics website.

We evaluate the clustering performance by two criteria: normalized mutual information (NMI) and adjusted Rand index (ARI) \citep{Man08}. NMI and ARI require the known ground-truth labels of cells, and we use the cell type labels of gene expression data provided in the secondary analysis outputs at the 10X Genomics website as the ground-truth labels.  Assume that $G$ is the known cell type labels and $P$ is the predicted clustering assignments, we then calculate NMI by
$
\frac{I(G;P)}{\sqrt{E(G) \times E(P)}},
$
where $I(G;P)$ represents the mutual information over $G$ and $P$, and $E(\cdot)$ represents the entropy. Assume that $n$ is the total number of single cells,
$n_{P,i}$ is the number of cells assigned to the $i$-th cluster in $P$, $n_{G,j}$ is the number of cells
belonging to the $j$-th cell type in $G$, and $n_{i,j}$ is the number of overlapping cells between the $i$-th cluster in $P$ and the $j$-th cell type in $G$. As a corrected-for-chance version of the Rand index, ARI is calculated by
$
\frac{\sum_{ij}\binom{n_{i,j}}{2}-\big[\sum_{i}\binom{n_{P,i}}{2}\sum_{j}\binom{n_{G,j}}{2}\big/\binom{n}{2}\big]}{\frac{1}{2}\big[\sum_{i}
\binom{n_{P,i}}{2}+\sum_{j}\binom{n_{G,j}}{2}\big]-\big[\sum_{i}\binom{n_{P,i}}{2}\sum_{j}\binom{n_{G,j}}{2}\big]/\binom{n}{2}}.
$
Higher values of NMI and ARI indicate better clustering performance.

\subsection{Results}
Table~\ref{tab:real} shows the results of clustering the cells in four real datasets. scICML performs best in both ARI and NMI in datasets 1-3. In dataset 4, scAI has the best clustering performance in ARI and NMI, while scICML ranks the second in ARI and NMI. scICML has the better clustering performance compared with scICML$_{0}$. It suggests that incorporating the matching of the clusters of linked features across the gene activity score in scATAC-seq data and gene expression in scRNA-seq data is helpful in clustering the cells. This is because the matching of the aggregating pattern of linked features across data types considers the biological dependency across different types of genomic features.

\begin{figure}[ht!]
\centering
\includegraphics[width=1\textwidth]{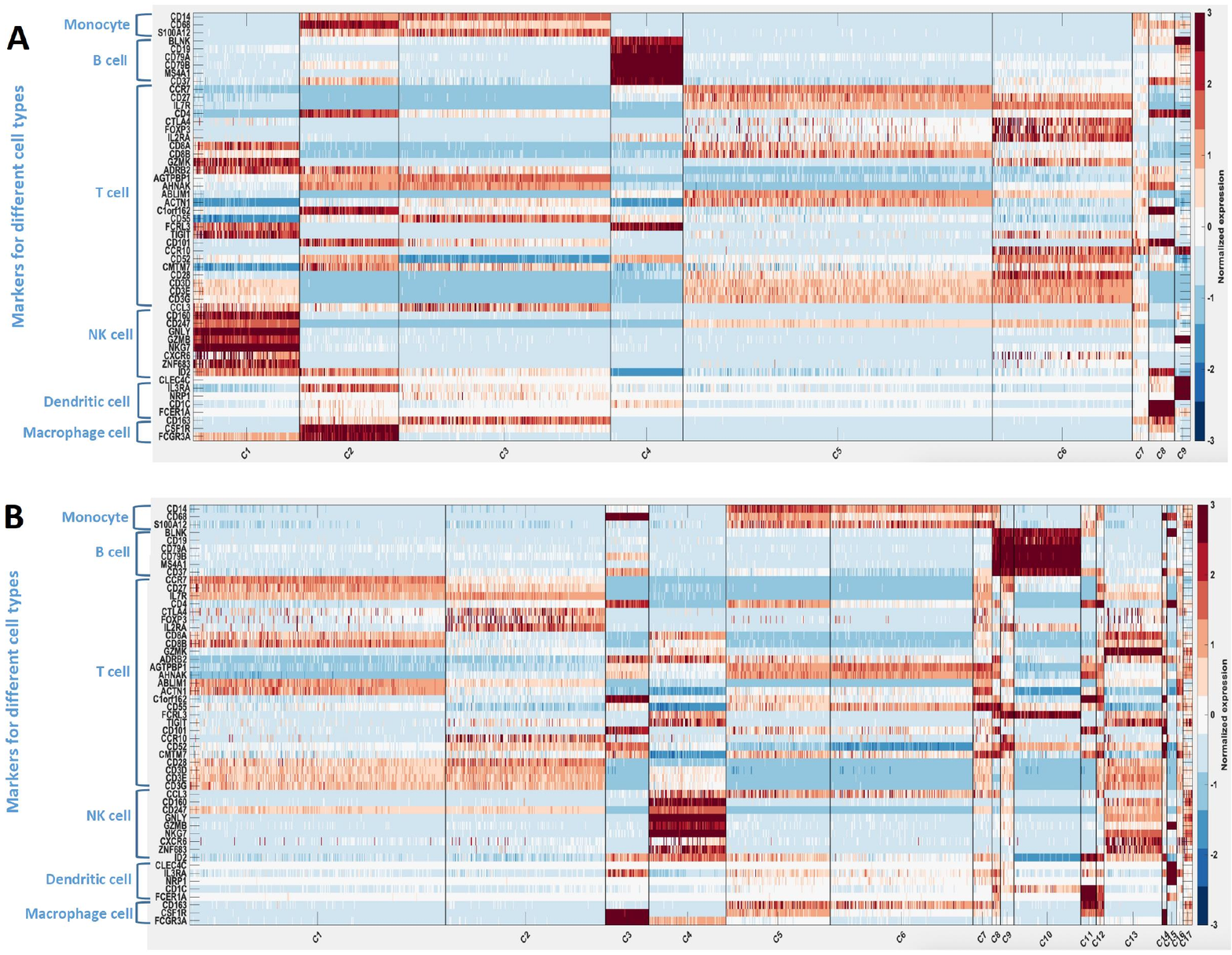}
\vspace{-9.5cm}
\caption{Identifying the cell types of the peripheral blood mononuclear cells by scICML. The heatmaps in (A) and (B) show the expression of the top 53 markers genes in clusters by scICML in datasets 1 and 2, respectively. We choose the specific markers for different cell types in peripheral blood mononuclear cells from CellMarker database \cite{Zhang2019}. For better visualization, we randomly average every 20 cells within the same cell cluster to generate pseudocells for every heatmap.}\label{fig:celltypes}
\end{figure}

Figures~\ref{fig:celltypes}A and B display the expression of the top 53 markers of peripheral blood mononuclear cells (PBMCs) in datasets 1 and 2, respectively. Lymphocytes, including T cells, B cells and natural killer (NK) cells, make up the majority of the PBMC population, followed by monocytes, and only a small percentage of dendritic cells and macrophage cells \cite{Cham2015}. We choose the markers for these cell subpopulations from the CellMarker database \cite{Zhang2019}. First, Figure~\ref{fig:celltypes}A shows that the cluster C4 likely corresponds to B cells, while the clusters C5 and C6 likely correspond to T cells. Figure~\ref{fig:celltypes}B shows that the clusters C1 and C2 likely correspond to T cells, while the clusters C8, C9 and C10 likely correspond to B cells. Second, the cluster C1 in Figure~\ref{fig:celltypes}A and the clusters C4 and C13 in Figure~\ref{fig:celltypes}B are more like NK cells, but they are mixed with a subset of markers for T cells. This is likely due to the mixing of natural killer T (NKT) cells in these clusters. NKT cells are a heterogeneous group of T cells that share properties of both T cells and NK cells \cite{Jerud2006}. Third, we notice that the cell subpopulations monocytes, a subset of T cells, dendritic cells and macrophage cells are mixed together, as shown in the clusters C2 and C3 in Figure~\ref{fig:celltypes}A, and the clusters C5, C6 and C7 in Figure~\ref{fig:celltypes}B. This may be caused by the fact that the monocytes are in the stage of differentiation into dendritic cells and macrophage cells \cite{Yang2014}, or due to the occurrence of doublets, i.e., the mixing of two cells of different types (say, T cell and macrophage cell) in one droplet while doing the droplet-based single-cell sequencing. 

\begin{table}[ht!]
\centering
\caption{The results of clustering the cells in four real datasets, evaluated by NMI and ARI. NMI and ARI are computed by the cell type labels provided by the analysis of gene expression data at the 10X Genomics website. The bold numbers represent the best clustering results.} \label{tab:real}
\resizebox{\textwidth}{!}{%
\begin{tabular}{cccccccccccccc}
\hline
&&&\multicolumn{2}{c}{Dataset 1}&&\multicolumn{2}{c}{Dataset 2}&&\multicolumn{2}{c}{Dataset 3}&&\multicolumn{2}{c}{Dataset 4}\\
&Clustering methods &&\multicolumn{2}{c}{($n=2,711$)}&&\multicolumn{2}{c}{($n=11,898$)}&&\multicolumn{2}{c}{($n=3,233$)}&&\multicolumn{2}{c}{($n=4,878$)}\\
\cline{4-5}\cline{7-8}\cline{10-11}\cline{13-14}
& &&NMI&ARI&&NMI&ARI&&NMI&ARI&&NMI&ARI\\
\hline
&scAI                          &&0.65&0.47&   &0.62&0.37&  &0.64&$0.44$    && $\bs{0.57}$&$\bs{0.47}$\\
&MOFA+                         &&0.60&0.43&   &0.58&0.39&  &0.54&0.33      && 0.53&0.38\\
&Seurat V4                     &&0.60&0.39&   &0.60&0.43&  &0.62&0.40      && 0.56&0.36\\
&scICML$_{0}$                    &&0.64&0.47&   &0.61&0.40&  &0.61&0.37      && 0.53&0.39\\
&scICML                          &&$\bs{0.67}$&$\bs{0.49}$&   &$\bs{0.64}$&$\bs{0.45}$&  &$\bs{0.69}$&$\bs{0.49}$      && 0.56&0.41\\

\hline
\end{tabular}%
}
\end{table}

\subsection{Comparison of running time}
\label{tm}
We summarize the computational cost by our method scICML, scAI, MOFA+ and Seruat V4 in Table \ref{tab:time}. The running time for scICML is less than MOFA+ for all the datasets. scAI runs faster than our method scICML for smaller datasets ($<5,000$ cells), but it takes much more time than scICML for larger dataset ($>10,000$ cells). Seurat V4 runs the fastest among all the methods.

\begin{table}[ht!]
\centering
\caption{The computational time by four clustering methods for four real-world datasets. All these algorithms are run on iMac platform with processor - 3.4 GHz Intel Core i5 and memory - 8 GB 2400 MHz DDR4. Both scICML and scAI are implemented on MATLAB (version R2020b), and both MOFA+ and Seurat V4 are implemented on RSdudio (version 2021.09.1).} \label{tab:time}
\resizebox{\textwidth}{!}{%
\begin{tabular}{cccccccccc}
\hline
&&&Dataset 1&&Dataset 2&&Dataset 3&&Dataset 4\\
&Clustering methods &&($n=2,711$)&&($n=11,898$)&&($n=3,233$)&&($n=4,878$)\\
\hline
&scAI             &&5.30(min)&   &600.70(min)&  &6.93(min) && 20.09(min)\\
&MOFA+            &&18.65(min)&  &145.44(min)&  &56.20(min) && 58.45(min)\\
&Seurat V4        &&1.90(min)&   &3.96(min)&    &2.37(min)  && 2.73(min)\\
&scICML             &&15.83(min)&   &112.5(min)&    &17.62(min)  && 24.75(min)\\
\hline
\end{tabular}%
}
\end{table}

\section{Discussion}
We have proposed scICML, a novel information-theoretic co-clustering-based multi-view learning method for the integrative analysis of single-cell multi-omics data. scICML performs four co-clusterings simultaneously on multiple views of single-cell data and matches the similar feature clusters for the linked features across different feature types. First, it uncovers the same cell clusters across the four views of data and aggregates similar features for each view of data, which is helpful to reduce the noise in high-dimensional single-cell genomic data. Second, it takes into account the biological dependency across different types of genomic features through matching the clusters for the linked features across data types. We have evaluated scICML on four different real-world single-cell multi-omics datasets, which demonstrate that scICML improves the overall clustering performance and provides biologically meaningful clustering results for well-studied data such as peripheral blood mononuclear cells. We have also shown the computational cost of scICML, and it has comparable computational speed in clustering the multi-omics datasets. Further improvement in computational speed may be achieved by developing mini-batch version of the algorithm. scICML has a broader potential applications. Take the data integration of scRNA-seq data and sc-methylation data from the same cells for example. Gene body methylation in sc-methylation data are directly linked with gene expression in scRNA-seq data, and this setting applies to our method scICML. In the near future, we will also study the data integration for different bio-molecules at the genome, transcriptome, translatome and proteome scales.

\bibliography{icml}

\begin{thebibliography}{}

\bibitem[Angermueller et~al., 2016]{Angermueller2016}
Angermueller, C., Clark, S.~J., Lee, H.~J., and the others (2016).
\newblock Parallel single-cell sequencing links transcriptional and epigenetic
  heterogeneity.
\newblock {\em Nature Methods}, 13:229--232.

\bibitem[Argelaguet et~al., 2020]{Arge2020}
Argelaguet, R., Arnol, D., Bredikhin, D., and so~on (2020).
\newblock Mofa+: a statistical framework for comprehensive integration of
  multi-modal single-cell data.
\newblock {\em Genome Biol}, 21(111).

\bibitem[Argelaguet et~al., 2018]{Arge2018}
Argelaguet, R., Velten, B., Arnol, D., Dietrich, S., Marioni, J.~C., and so~on
  (2018).
\newblock Multi-omics factor analysis-a framework for unsupervised integration
  of multi-omics data sets.
\newblock {\em Mol Syst Biol}, 14.

\bibitem[Butler et~al., 2018]{Butler2018}
Butler, A., Hoffman, P., Smibert, P., Papalexi, E., and Satijia, R. (2018).
\newblock Integrating single-cell transcriptomic data across different
  conditions, technologies, and species.
\newblock {\em Nat Biotechnol}, 36(5):411--420.

\bibitem[Calinski and Harabasz, 1974]{Calin1974}
Calinski, R.~B. and Harabasz, J. (1974).
\newblock A dendrite method for cluster analysis.
\newblock {\em Communications in Statistics}, 3:1--27.

\bibitem[Cao et~al., 2018]{Cao2018}
Cao, J., Cusanovich, D.~A., Ramani, V., and the others (2018).
\newblock Joint profiling of chromatin accessibility and gene expression in
  thousands of single cells.
\newblock {\em Science}, 361(6409):1380--1385.

\bibitem[Chen et~al., 2019a]{chen2019asse}
Chen, H., Lareau, C., Andreani, T., and the others (2019a).
\newblock Assessment of computational methods for the analysis of single-cell
  atac-seq data.
\newblock {\em Genome biology}, 20(1):1--25.

\bibitem[Chen et~al., 2019b]{Chen2019}
Chen, S., Lake, B.~B., and Zhang, K. (2019b).
\newblock High-throughput sequencing of the transcriptome and chromatin
  accessibility in the same cell.
\newblock {\em Nature Biotechnology}, 37:1452--1457.

\bibitem[Chen et~al., 2020]{Ma2020}
Chen, S., Lake, B.~B., and Zhang, K. (2020).
\newblock Chromatin potential identified by shared single-cell profiling of rna
  and chromatin.
\newblock {\em Cell}, 183:1103--1116.

\bibitem[Christopher et~al., 2008]{Man08}
Christopher, D.~M., Prabhakar, R., and Hinrich, S. (2008).
\newblock {\em Introduction to Information Retrieval}.
\newblock Cambridge University Press.

\bibitem[Clark et~al., 2018]{Clark2018}
Clark, S.~J., Argelaguet, R., Kapourani, C.~A., and the others (2018).
\newblock scnmt-seq enables joint profiling of chromatin accessibility, dna
  methylation and transcription in single cells.
\newblock {\em Nature Communications}, 9(781).

\bibitem[Colome-Tatche and Theis, 2018]{CT2018}
Colome-Tatche, M. and Theis, F.~J. (2018).
\newblock Statistical single cell multi-omics integration.
\newblock {\em Curr Opin Syst Biol}, 7:54--9.

\bibitem[Cover and Thomas, 1991]{Cover1991}
Cover, T.~M. and Thomas, J.~A. (1991).
\newblock Elements of information theory.
\newblock {\em Wiley-Interscience}.

\bibitem[Dai et~al., 2008]{Dai2008}
Dai, W.~Y., Yang, Q., Xue, G.~R., and Yu, Y. (2008).
\newblock Self-taught clustering.
\newblock {\em Proceedings of the 25th international Conference on Machine
  Learning}.

\bibitem[David et~al., 2020]{David2020}
David, L., Johannes, K., Ewa, S., and the others (2020).
\newblock Eleven grand challenges in single-cell data science.
\newblock {\em Genome Biol}, 21(31).

\bibitem[Dhillon et~al., 2003]{Dhillon2003}
Dhillon, I.~S., Mallela, S., and Modha, D.~S. (2003).
\newblock Information-theoretic co-clustering.
\newblock {\em Proceedings of the Ninth ACM SIGKDD International Conference on
  Knowledge Discovery and Data Mining}, pages 89--98.

\bibitem[Duren et~al., 2018]{Duren2018}
Duren, Z., Chen, X., Zamanighomi, M., Zeng, W., Satpathy, A., Chang, H., Wang,
  Y., and Wong, W.~H. (2018).
\newblock Integrative analysis of single cell genomics data by coupled
  non-negative matrix factorizations.
\newblock {\em Proc. Natl. Acad. Sci.}, (115):7723--7728.

\bibitem[Hao et~al., 2021]{Hao2021}
Hao, Y., Hao, S., Andersen-Nissen, E., Mauck, W.~M., Zheng, S., and butler, A.
  (2021).
\newblock Integrated analysis of multimodal single-cell data.
\newblock {\em Cell}, 184(13):3573--3587.

\bibitem[Jerud et~al., 2006]{Jerud2006}
Jerud, E.~S., Bricard, G., and Porcelli, S.~A. (2006).
\newblock Natural killer t cells: Roles in tumor immunosurveillance and
  tolerance.
\newblock {\em Transfus. Med. Hemother.}, 33(1):18--36.

\bibitem[Jin et~al., 2020]{Jin2020}
Jin, S., Zhang, L., and Nie, Q. (2020).
\newblock scai: an unsupervised approach for the integrative analysis of
  parallel single-cell transcriptomic and epigenomic profiles.
\newblock {\em Genome Biology}, 21(25).

\bibitem[Lin et~al., 2019]{Lin2019}
Lin, Z.~X., Zamanighomi, M., Daley, T., Ma, S., and Wong, W.~H. (2019).
\newblock Model-based approach to the joint analysis of single-cell data on
  chromatin accessibility and gene expression.
\newblock {\em Stat. Sci}.

\bibitem[Liu et~al., 2019]{Liu2019}
Liu, L., Liu, C., Quintero, A., and the others (2019).
\newblock Deconvolution of single-cell multi-omics layers reveals regulatory
  heterogeneity.
\newblock {\em Nature Commnunications}, 10(470).

\bibitem[Ma et~al., 2022]{Ma2022}
Ma, Y., Sun, Z., Zeng, P., Zhang, W., and Lin, Z. (2022).
\newblock Jsnmf enables effective and accurate integrative analysis of
  single-cell multiomics data.
\newblock {\em Briefings in Bioinformatics}.

\bibitem[Macaulay et~al., 2017]{Macau2017}
Macaulay, I.~C., Ponting, C.~P., and Voet, T. (2017).
\newblock Single-cell multiomics: multiple measurements from single cells.
\newblock {\em Trends Genet.}, 33:115--68.

\bibitem[Stuart et~al., 2019]{Stuart2019}
Stuart, T., Butler, A., Hoffman, P., and the others (2019).
\newblock Comprehensive integration of single-cell data.
\newblock {\em Cell}, (177):1888--1902.

\bibitem[Verhoeckx et~al., 2015]{Cham2015}
Verhoeckx, K., Cotter, P., Lopez-Exposit, I., Kleiveland, C., Lea, T., Mackie,
  A., et~al. (2015).
\newblock {\em The impact of food bioactives on health: in vitro and ex vivo
  models [Internet]}.
\newblock Cham (CH): Springer.

\bibitem[Wangwu et~al., 2021]{Wangwu2021}
Wangwu, J., Sun, Z., and Lin, Z. (2021).
\newblock scamace: model-based approach to the joint analysis of single-cell
  data on chromatin accessibility, gene expression and methylation.
\newblock {\em Bioinformatics}, 37(21):3874--3880.

\bibitem[Welch et~al., 2019]{Welch2019}
Welch, J.~D., Kozareva, V., Ferreira, A., Vanderburg, C., and the others
  (2019).
\newblock Single-cell multi-omic integration compares and contrasts features of
  brain cell identity.
\newblock {\em Cell}, 177(7):1873--1887.

\bibitem[Yang et~al., 2014]{Yang2014}
Yang, J., Zhang, L., Yu, C., Yang, X.~F., and Wang, H. (2014).
\newblock Monocyte and macrohage differentiation: circulation inflammatory
  monocyte as biomarker for inflammatory diseases.
\newblock {\em Biomarker Research}, 2(1).

\bibitem[Zeng and Lin, 2020]{Zeng2020}
Zeng, P. and Lin, Z. (2020).
\newblock Coupled co-clustering-based unsupervised transfer learning for the
  ingetrative analysis of single-cell genomics data.
\newblock {\em Briefings in bioinformatics.}

\bibitem[Zeng and Lin, 2021]{Zeng2021}
Zeng, P. and Lin, Z. (2021).
\newblock couplecoc+: An information-theoretic co-clustering-based transfer
  learning framework for the integrative analysis of single-cell genomic data.
\newblock {\em PLoS Comput Biol}, 17(6).

\bibitem[Zhang et~al., 2018]{Zhang2018}
Zhang, H., Lee, C.~A.~A., Li, Z., and the others (2018).
\newblock A multitask clustering approach for single-cell rna-seq analysis in
  recessive dystrophic epidermolysis bullosa.
\newblock {\em PLoS Comput Biol}, 14(4).

\bibitem[Zhang and Nie, 2021]{Zhang2021}
Zhang, L. and Nie, Q. (2021).
\newblock scmc learns biological variation through the alignment of multiple
  single-cell genomics datasets.
\newblock {\em Genome Biology}, 22(10).

\bibitem[Zhang et~al., 2019]{Zhang2019}
Zhang, X., Lan, Y., Xu, J., Quan, F., Zhao, E., Deng, C., et~al. (2019).
\newblock Cellmarkers: a manually curated resource of cell markers in human and
  mouse.
\newblock {\em Nucleic Acids Res.}, 47:721--728.

\bibitem[Zhu et~al., 2019]{Zhu2019}
Zhu, C., Yu, M., Huang, H., and the others (2019).
\newblock An ultra high-throughput method for single-cell joint analysis of
  open chromatin and transcriptome.
\newblock {\em Nature Structural \& Molecular Biology}, 26:1063--1070.

\bibitem[Zhu et~al., 2021]{Zhu2021}
Zhu, C., Zhang, Y., Li, Y.~E., and the others (2021).
\newblock Joint profiling of histone modifications and transcriptome in single
  cells from mouse brain.
\newblock {\em Nature Methods}, 18:283--292.

\end{thebibliography}
\bibliographystyle{apalike}

\end{document}